\PassOptionsToPackage{dvipsnames}{xcolor} 
\documentclass{llncs}
\usepackage{graphicx} 
\usepackage[dvipsnames]{xcolor}
\usepackage{amsmath, amssymb}   
\usepackage{graphics}
\usepackage{wrapfig}
\usepackage{caption}
\usepackage{verbatim}
\usepackage{paralist}
\usepackage{cancel}
\usepackage{xspace}
\usepackage{multicol}
\usepackage{subcaption}
\usepackage{algorithm}
\usepackage{algpseudocode}
\usepackage{tabularx}
\usepackage{diagbox}
\usepackage{cite}

\usepackage{tikz,ifthen,pgfplots}
\usetikzlibrary{arrows,trees,backgrounds,automata,shapes,decorations,plotmarks,fit,calc,positioning,shadows,chains}
\tikzstyle{every pin edge}=[<-,shorten <=1pt]
\tikzstyle{neuron}=[circle,fill=black!25,minimum size=17pt,inner sep=0pt]
\tikzstyle{input neuron}=[neuron, fill=green!40]
\tikzstyle{output neuron}=[neuron, fill=red!40]
\tikzstyle{hidden neuron}=[neuron, fill=blue!40]
\tikzstyle{constructed neuron}=[neuron, fill=orange!50]
\tikzstyle{annot} = [text width=6em, text centered]
\tikzstyle{nnedge} = [-{stealth},shorten >=0.1cm, shorten <=0.05cm,line width=0.8pt,black]
\usetikzlibrary{calc}

\usepackage[numbers,sectionbib]{natbib}
\newcommand{\image}{x'}
\newcommand{\sat}{\texttt{SAT}}
\newcommand{\unsat}{\texttt{UNSAT}}
\newcommand{\unknown}{\texttt{UNKNOWN}}
\newcommand{\timeout}{\texttt{TIMEOUT}}
\newcommand{\rn}[1]{\mathbb{R}^{#1}}
\newcommand{\nn}{\mathcal{N}}
\newcommand{\mysubsection}[1]{\medskip\noindent\textbf{#1}}
\newcommand{\allones}[1]{J_{#1}}
\newcommand{\relu}{\text{ReLU}\xspace{}}

\title{Robustness Assessment of a Runway Object Classifier for Safe Aircraft Taxiing}
\author{
    Yizhak Elboher\inst{1}\thanks{Authors contributed equally.} \and
    Raya Elsaleh\inst{1}$^\star$ \and
    Omri Isac\inst{1}$^\star$ \and
    M\'elanie Ducoffe\inst{2} \and
    Audrey Galametz\inst{2} \and
    Guillaume Pov\'eda\inst{2} \and
    Ryma Boumazouza\inst{2} \and
    No\'emie Cohen\inst{2} \and
    Guy Katz\inst{1}
}
\institute{ 
The Hebrew University of Jerusalem
\and Airbus Central Research \& Technology, AI Research\\
\email{\{yizhak.elboher, raya.elsaleh, omri.isac, guy.katz\}@mail.huji.ac.il \\
\{melanie.ducoffe, audrey.galametz, guillaume.poveda, ryma.boumazouza,\\noemie-lucie-virginie.cohen\}@airbus.com
}
\authorrunning{Y. Elboher, R. Elsaleh, O. Isac  et al.}}

\begin{document}
\maketitle
\setcounter{footnote}{0} 
\begin{abstract}
  As deep neural networks (DNNs) are becoming the prominent solution
  for many computational problems, the aviation industry seeks to
  explore their potential in alleviating pilot workload and
  improving operational safety. However, the use of DNNs in these types
  of safety-critical applications requires a thorough certification
  process. This need could be partially addressed through formal verification,
  which provides rigorous assurances --- e.g.,~by proving the absence
  of certain mispredictions. In this case-study paper, we demonstrate
  this process on an image-classifier DNN currently under
  development at Airbus, which is intended for use during the aircraft
  taxiing phase. We use formal methods to assess this DNN's robustness
  to three common image perturbation types: \emph{noise},
  \emph{brightness} and \emph{contrast}, and some of their
  combinations. This process entails multiple invocations of the
  underlying verifier, which might be computationally expensive; and
  we therefore propose a method that leverages the monotonicity of
  these robustness properties, as well as the results of past
  verification queries, in order to reduce the overall number of
  verification queries required by nearly 60\%. Our results indicate the level of robustness achieved by the DNN classifier
  under study, and indicate that it is considerably more vulnerable to
  noise than to brightness or contrast
  perturbations.
\end{abstract}

\section{Introduction}
\label{sec:intro}

In recent years, deep neural networks (DNNs) have been revolutionizing
computer science, advancing the state of the art in many
domains~\cite{SzToEr13} --- including natural language processing,
computer vision, and many others. In the aviation domain, aircraft
manufacturers are now exploring how deep-learning-based technologies
could decrease the cognitive load on pilots, while increasing the
safety and operational efficiency of, e.g., airports. In particular,
these technologies could prove useful during the aircraft taxi phase,
which often creates an increased cognitive load on pilots who have to
simultaneously manage the flight plan, the aircraft itself, and any
objects on the tarmac.

Despite their success, DNNs are known to be prone to various
errors. Notable among these are \emph{adversarial
  inputs}~\cite{GoShSz14}, which are slightly perturbed inputs that
lead to incorrect and potentially unsafe DNN outputs.
While there exist many techniques for efficiently finding adversarial
inputs, it is unclear how to certify that no
such examples exist. However, such a certification process will be
required to allow the integration of DNNs into safety-critical
industrial systems, e.g., in aviation.

Aviation authorities involved in managing aircraft certification,
  such as the European Union Aviation Safety Agency (EASA), have
  recently published the key elements required for certifying DNN
  models to be used in
  aviation.\footnote{https://www.easa.europa.eu/en/downloads/137631/en}
  There, EASA particularly emphasizes that DNN verification solutions,
  to be applied during the learning and system integration phases, will likely
  constitute a \emph{means of
  compliance} with regulatory requirements.\footnote{https://www.easa.europa.eu/en/document-library/general-publications/concepts-design-assurance-neural-networks-codann}
EASA points out, however, that the current scalability and
the expressiveness of DNN verification techniques is limited.




Typically, DNN formal verification tools seek to prove that, for a
given infinite set of inputs, a DNN only produces outputs that fall
within a safe subspace of the output space.   To date, these tools have
  been predominantly applied in assessing the robustness of DNN
  predictions against specific types of local input
  perturbations. Maturing these techniques is thus key in allowing them to
  meet the bar needed for DNN certification in, e.g., aviation. 
  This point is again stressed in EASA's AI
  Roadmap,\footnote{https://www.easa.europa.eu/en/domains/research-innovation/ai}
  which emphasizes the need for providing more general guarantees of a DNN's
  stability.

Although DNN verification has been making great
strides~\cite{BrMuBaJoLi23, HeLo20, GoKaPaBa18, MaNoOr19, WuZeKaBa22,
  OsBaKa22, GaGePuVe19, HuKwWaWu17, MuMaSiPuVe22, LyKoKoWoLiDa20}, it
has so far been applied only to a limited number of real-world systems. In this case-study paper, we study the applicability and scalability of DNN verification through an object classification use-case, relevant to the aviation domain and of specific interest to Airbus. We explore pertinent vision-oriented perturbations (\emph{noise}, \emph{brightness}, and \emph{contrast}) and use formal verification to quantify their effects on DNN's robustness. As a
back-end engine, we use the Marabou DNN
verifier~\cite{WuIsZeTaDaKoReAmJuBaHuLaWuZhKoKaBa24}. We also demonstrate that
the verification process can be optimized by
leveraging the monotonicity of the studied perturbations.

Our results indicate that while the DNN is highly 
sensitive to noise perturbations, it is slightly less vulnerable to
contrast and brightness perturbations. This is a reassuring result, as
these perturbations are strongly correlated with highly unpredictable
operating conditions, especially outdoors. 
More broadly, our results showcase the usefulness and
potential of DNN verification in aviation that could easily be extended to other safety-critical domains.


\section{Background}
\label{sec:background}
\label{sec:DNNs}


\noindent
\textbf{Deep Neural Networks.}
A deep neural network~\cite{GoBeCu16} $ \nn:\rn{n}\rightarrow\rn{k} $
is comprised of $ m $ layers, $ L_1,...,L_m $. Each layer $L_i$
consists of a set of nodes, $S_i$. When $\nn$ is evaluated, each node
in the input layer is assigned an initial value. Then, the value of the $j^{th}$ node in the
$ 2 \leq i < m $ layer, $v_j^i$, is computed as:
\begin{center}
	$v_j^i = f \left( \underset{l=1} { \overset {|S_{i-1}|} { \sum } }
	w^{i-1}_{j,l} \cdot v^{i-1}_l + b^i_j \right)$
\end{center}
where $f:\mathbb{R}\rightarrow\mathbb{R}$ is an \textit{activation
  function} and $w^{i-1}_{j,l}, b^i_j \in \mathbb{R}$ are the
respective \emph{weights} and \emph{biases} of $\nn$. The most common
activation function is the \textit{rectified linear unit (ReLU)}, defined as
$\relu(x) = max(0,x)$.  Finally, neurons in the output layer are
assigned values using an affine combination only. The output of the
DNN is the values of the nodes in its final layer.  An
image-classifier $\nn: \rn{n}\rightarrow C\subset\mathbb{N}$
assigns each input image $\image$ a class $c\in C$, which describes
the main object depicted in $\image$. For convenience, $\image$ is
regarded as both a vector and a matrix, interchangeably. For an example of
a DNN and its evaluation, see Appendix~\ref{app:DNNS}.

\mysubsection{DNN Verification.}
For a DNN $\nn:\rn{n}\rightarrow\rn{k} $, input property
$ P \subset \rn{n}$ and output property $Q \subset \rn{k} $, the
\emph{DNN verification problem} is to decide whether there exist
$ x \in P$ and $y\in Q$ such that $ \nn(x) = y $.  If such a pair
exists, the verification query $(\mathcal{N},P, Q)$ is
\textit{satisfiable} (\sat), and the pair $(x,y)$ is called a witness;
otherwise, it is \textit{unsatisfiable} (\unsat). Typically, $Q$
encodes an undesired behavior, and so a witness is a
\emph{counterexample} that demonstrates an error.

\section{Industrial Use-Case: Runway Object Classification}
\label{sec:usecase}

\subsection{Runway Object Classification}  

In 2020, Airbus concluded its Autonomous Taxi, Take-Off and Landing
(ATTOL)
project.\footnote{https://www.airbus.com/en/newsroom/press-releases/2020-06-airbus-concludes-attol-with-fully-autonomous-flight-tests}
The objective of ATTOL was to design a fully
autonomous controller for the taxi, take-off, approach and landing
phases of a commercial aircraft --- by leveraging state-of-the-art
technology, and in particular deep-learning models used for
vision-assisted functions. As part of the project, $400$ flights over
a period of two years were instrumented to collect video data from
aircraft in operation. This unique dataset is currently being used to
further mature several vision-based functions within Airbus. Using
this dataset, it was observed that the taxi phase of the flight, in
particular, could benefit from autonomous support. During this phase,
pilots are conducting aircraft operations, while simultaneously
dealing with the unpredictable nature of airport management and
traffic. Object identification, in particular of potential threats on
the runway, could thus support the pilots during this phase. Several object classifiers are being tested for this purpose within Airbus.

In this study, we focus on images of runway objects extracted from
taxiing videos --- i.e.,~all objects are observed from an aircraft on the ground. 
We extract $(224 \times 224)$ pixel images from the original, high-resolution gray-scaled images, centered on a specific runway object. A DNN $N_1$ is trained on resampled $(32 \times 32)$ images. The four considered classes are \emph{Aircraft}, \emph{Vehicle}, \emph{Person}, and
\emph{Negative}, extracted where no object is found.  $N_1$ is a
feedforward DNN, with roughly $8000$ \relu{} neurons, and an accuracy of $85.3\%$ on the test dataset ($1145 / 1342$ images).\footnote{These DNNs will not be used as such in Airbus products. More robust models are currently under development, in part supported by analyses such as the one presented here.}

\subsection{Properties of Interest}  

We seek to verify the \emph{local robustness} of a runway object
classifier $\nn$; i.e.,~that small perturbations around a correctly-classified input $\image$ do not cause misclassification, encoded by $Q$.
We specify $Q$ as:
$Q_{\image} := C \setminus \nn(\image)$. We use the input property $P$ to define three perturbation types: noise, brightness, and contrast.

\mysubsection{Noise}.  In this widely studied form of
perturbation~\cite{KaBaDiJuKo21, CaKoDaKoKaAmRe22},
the perturbed input images are taken from an $\epsilon$-ball around
$\image$: $P=B_\epsilon (\image)$, where
$B_\epsilon$ is the $\ell_\infty$-$\epsilon$-ball around $\image$, and
$\epsilon>0$.

\mysubsection{Brightness.} A brightness perturbation is caused by 
shifting  all pixels of $\image$ by a constant value $b$:
$bright(\image,b) := \image + b \cdot \allones{n}$, where
$\allones{n}$ is the all-ones matrix of size $n\times n$.  We define
$P=bright_\beta(\image) := \lbrace bright(\image,b) \mid |b|\leq \beta
\rbrace$ for some $\beta>0$, to allow all brightness perturbations of
absolute value at most $\beta$.  See Appendix~\ref{app:perturbations}
for a visual example.

\mysubsection{Contrast.} A contrast perturbation $con(\image, c,\mu)$
is created by scaling all image pixels multiplicatively, rescaling
their difference from a mean value $\mu \in [0,1]$ by a multiplicative
constant $c\in\mathbb{R}_{\geq0}$:
$con(\image, c,\mu) := \mu \cdot \allones{n} + c\cdot(\image -
\mu\cdot \allones{n}) $.  We then set
$P=con_{\gamma,\mu}(\image) := \lbrace con(\image, c ,\mu) \mid |c -
1| \leq \gamma \rbrace$, to encode all contrast perturbations with
value of at most $\gamma$, where $\mu$ remains constant and
$\gamma\in[0,1]$.  See Appendix~\ref{app:perturbations} for a visual
example.

\section{The Formal Verification Process}
\label{sec:contribution}

\subsection{Encoding Brightness and Contrast Perturbations}
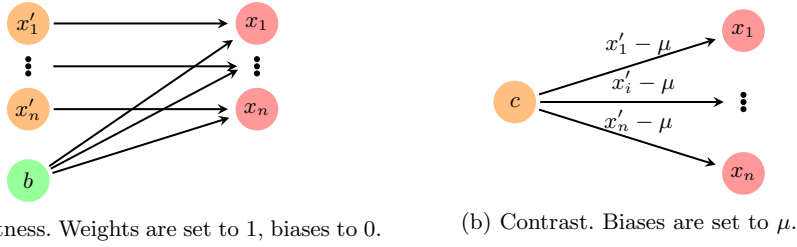
\begin{figure}[h!]
\centering
\def\layersep{1.6cm}
\scalebox{0.95}{
 \begin{tabular}[]{cc}
	\begin{subfigure}[t!]{0.55\linewidth}
 
        \begin{center}
        \begin{tikzpicture}[shorten >=1pt,->,draw=black!50, node distance=\layersep,font=\footnotesize]
                
            \node[constructed neuron] (I-1) at (0,0) {$x'_1$};
            \filldraw [black] (0,-0.5) circle (1pt);
            \filldraw [black] (0,-0.6) circle (1pt);
            \filldraw [black] (0,-0.7) circle (1pt);
            \node[constructed neuron] (I-2) at (0,-1.2) {$x'_n$};
            \node[input neuron] (B) at (0,-2.2) {$b$};

            \node[output neuron] (C-1) at (2*\layersep,0) {$x_1$};
            \filldraw [black] (2*\layersep,-0.5) circle (1pt);
            \filldraw [black] (2*\layersep,-0.6) circle (1pt);
            \filldraw [black] (2*\layersep,-0.7) circle (1pt);
            \node[output neuron] (C-2) at (2*\layersep,-1.2) {$x_n$};
            
            \draw[nnedge] (I-1) --node[above,pos=0.3] {} (C-1);
            \draw[nnedge] (0.3,-0.6) --node[above,pos=0.3] {} (2*\layersep-5,-0.6);
            \draw[nnedge] (I-2) --node[above,pos=0.3] {} (C-2);
            \draw[nnedge] (B) --node[above,pos=0.3] {} (C-1);
            \draw[nnedge] (B) --node[above,pos=0.3] {} (C-2);
            \draw[nnedge] (B) --node[above,pos=0.3] {} (2*\layersep-5,-0.6);

            \node[below=0.05cm of C-1] (b1) {};
                
        \end{tikzpicture}

        \caption{Brightness. Weights are set to $1$, biases to $0$.}
        \label{fig:brightness}
        \end{center}
    \end{subfigure}&

	\begin{subfigure}[t!]{0.55\linewidth}
         \begin{center}        
            \begin{tikzpicture}[shorten >=1pt,->,draw=black!50, node distance=\layersep,font=\footnotesize]
                
          \node[constructed neuron] (C) at (0,-1) {$c$};
            
            \node[output neuron] (C-1) at (2*\layersep,0) {$x_1$};
            \filldraw [black] (2*\layersep,-0.9) circle (1pt);
            \filldraw [black] (2*\layersep,-1) circle (1pt);
            \filldraw [black] (2*\layersep,-1.1) circle (1pt);
            \node[output neuron] (C-2) at (2*\layersep,-2) {$x_n$};

            \draw[nnedge] (C) --node[above,pos=0.55] {$\image_1-\mu$} (C-1);
            \draw[nnedge] (C) --node[above,pos=0.55] {$\image_i-\mu$} (2*\layersep-5,-1);
            \draw[nnedge] (C) --node[above,pos=0.55] {$\image_n-\mu$} (C-2);
                
            \node[below=0.05cm of C-1] (b1) {};
                
        \end{tikzpicture}
        \caption{Contrast. Biases are set to $\mu$.}
        \label{fig:contrast}
        \end{center}
    \end{subfigure}
    \end{tabular}
    }
    \caption{Modeling brightness and contrast perturbations by adding an input layer.}
    \label{fig:properties}
    \vspace{-0.4cm}
\end{figure}

We now show how to encode the brightness and contrast properties
described in Section~\ref{sec:background} into verification queries
that assess robustness to noise perturbations over a modified input
space. This reduction allows us to use any of the
available tools that support such queries as a backend. The encoding is performed
by adding a new input layer to the network, as illustrated in
Fig.~\ref{fig:properties}.

\mysubsection{Brightness}.  The new input layer clones the original
input layer and adds a single neuron $b$ to represent the brightness
perturbations. The weights from the new layer to the following,
original input layer are set to $1$ so that every variable
$x_i\in x$ is assigned $x_i = x'_i + b$. The bounds for the new neuron
are set to $b \leq \beta, b \geq -\beta$, whereas inputs $x_i'$ are exactly
restricted to the input around which the robustness is being
verified. We note that in this case, this single construction allows the verification of
the robustness around any input, by selecting appropriate $x_i'$
values. We further note that this construction can be used to
simultaneously encode noise and brightness perturbations, by bounding
the input neurons $x_i'$ to an $\epsilon$-ball around an input of
interest. This gives rise to two-dimensional queries, for any
combination of $\beta$ and $\epsilon$ values, which allows modeling a more
realistic nature of perturbations.

\mysubsection{Contrast.}  The new input layer contains a single input
neuron, $c$. We treat $\mu, \image$ as constants, and set the weights
from the new layer in a way that every neuron $x_i$ in that layer is
assigned $ x_i = (\image_i - \mu)\cdot c + \mu $. Finally, we set the
bounds $c \geq 1-\gamma, c \leq 1+\gamma$. We note that the contrast
perturbation is multiplicative with respect to $c$, $\mu$, and the
input image $\image$.  Since DNN verification algorithms typically
only support linear operations, either $\image$ or $c$ should be
fixed. Therefore, a separate DNN is constructed for each input image;
and there is no immediate way to encode a simultaneous noise
perturbation.

\subsection{Incremental Verification Algorithm}
\label{sec:incremental}
For any fixed image $\image$, we seek to solve numerous brightness,
noise and contrast robustness queries, with different values of
$\epsilon$, $\beta$ and $\gamma$. Since executing these queries is
computationally expensive, we exploit the monotonicity of these
properties to reduce their number. Let
$\beta' < \beta$, $\epsilon' < \epsilon$ and $\gamma' <
\gamma$. If there exists an adversarial example for parameters
$\beta'$, $\epsilon'$ or $\gamma'$, it then also constitutes a
counterexample for a query with parameters $\beta$, $\epsilon$ or
$\gamma$ respectively. Conversely, if the network is robust with
respect to parameters $\beta$, $\epsilon$ or $\gamma$, then it is also
robust to perturbation with parameters $\beta', \epsilon'$ or
$\gamma'$ respectively.

We exploit this property in a binary search algorithm for contrast
queries, and in our \emph{incremental verification algorithm} for
brightness and noise queries. Intuitively, the algorithm initializes a grid representing all combinations of $\epsilon, \beta$ parameters that need to be verified. 
The observation above states that for every row and every column, there is at most one transition from \unsat{} to \sat{}, which is represented by a step graph within the grid. The algorithm then discovers this step graph instead of solving all queries in the grid.


The pseudo-code of the algorithm appears in Algorithm~\ref{alg:inc}.
Formally, Algorithm~\ref{alg:inc} assumes the existence of a verification procedure $\text{verify}(\mathcal{N}, \image, \beta, \epsilon)$ which verifies the robustness of a network $\mathcal{N}$ to noise perturbations of value at most $\epsilon$ and brightness perturbations of value at most $\beta$ around an image $\image$.
The algorithm is given an input network $\mathcal{N}$, an image $\image$, and two increasingly ordered arrays $B,E$, containing the values of parameters $\beta$ and $\epsilon$ we intend to check, respectively. Then, the algorithm initializes a grid representing all possible combinations of parameters $(\beta,\epsilon)\in B\times E$, and a temporary tuple $(b,e)$ representing the lowest value of $\beta$ and highest value $\epsilon$ and corresponding to the top-left corner of the grid.
The algorithm then iteratively calls  $\text{verify}(\mathcal{N},
\image, b, e)$ to populate the grid. If the result is $\sat$, then for all queries with the same $\epsilon$ value, and a greater $\beta$ (all cells to the right of the current cell) the result is $\sat$ as well.\footnote{Note that this is the case for all queries with greater values of $\epsilon,\beta$ (the top right rectangle), though the values of queries with a greater $\epsilon$ value are already decided. A dual argument applies to the \unsat{} case as well.} We then mark the relevant cells with $\sat$ and decrement $\epsilon$ to the next value.
If the result is $\unsat$, then for all queries with the same $\beta$ value, and a smaller $\epsilon$ (all cells to the bottom of the current cell), the result is $\unsat$ as well. We then mark the relevant cells with $\unsat$ and increment $\beta$ to the next value.
When the current cell reaches the final column or row, a typical binary search algorithm is used to find the remaining results.
Note that the algorithm requires only $O(m)$ calls 
to the verifier while naively solving all queries requires $O(m^2)$ calls, where $m$ is the number of
possible values of $\beta$ or $\epsilon$ (the maximal of the two). 
For contrast queries, the binary search allows using a logarithmic number of invocations of the verifier instead of a linear number. 

To support the use of real-world verifiers, we also address
cases where the verifier returns a 
 \timeout{} or error value. When these cases occur, we mark the
corresponding cell  $(e, b)$ with an $\unknown$ result, and increment the value of $b$ as if the result was $\sat$.
In addition, we use binary search for the remaining values of $e$, where the value of $b$ is constant. Note that in the presence of \timeout, the bound of $O(m)$ calls to the verifier is not guaranteed.

\begin{algorithm}[t!]
\textbf{Input:} Arrays $B,E$ with values of $\epsilon,\beta$ in increasing order, respectively, a verifier V, a network $\mathcal{N}$ and an image $\image$. \\
\textbf{Output:} A grid representing the robustness of $\mathcal{N}$ to brightness and noise perturbations around $\image$, for all values in $B,E$.
\begin{algorithmic}
\State $b \gets 0$
\State $e \gets length(E) - 1$
\State {$grid \gets 0_{length(B) \times length(E)} $}
\While {$b < length(B)$ and $e \geq 0$}
    \If{ $b = length(B) - 1$}
        \State{Binary search with remaining values of $e$; $b$ is constant.}
    \EndIf
    \If{ $e = 0$} 
        \State{Binary search with remaining values of $b$; $e$ is constant.}
    \EndIf
    
    \State $result \gets V.\text{verify}(\mathcal{N}, \image, E[e], B[b])$
    \If{$result = \sat$}
        \State {$\forall i \geq b : grid[i][e] \gets \sat$}
        \State{$e \gets e - 1$}
    \ElsIf{$result = \unsat$}
            \State {$\forall j \leq e: grid[b][j] \gets \unsat$}
            \State{$b \gets b + 1$}
    \Else \Comment{Timeout, Memoryout, etc.}
    \State {$grid[b][e] \gets \unknown$}
        \State{Binary search with remaining values of $e$; $b$ is constant.}
        \State{$b \gets b + 1$}
    \EndIf
\EndWhile
\State {\textbf{return} grid}
\end{algorithmic}
    \caption{Incremental verification algorithm}
    \label{alg:inc}
\end{algorithm}

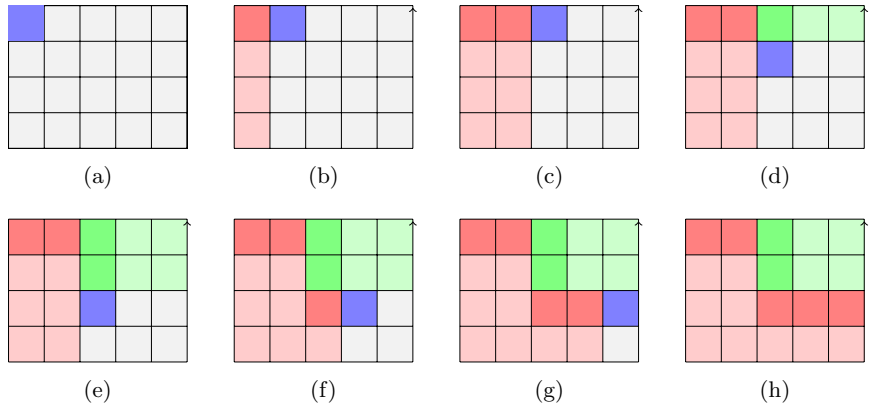
\begin{figure}[t!]
\scalebox{0.95}{
\begin{tabular}{cccc}
	\begin{subfigure}[t!]{0.25\textwidth}
    \centering
    \begin{tikzpicture}[shorten >=1pt,->,draw=black!50,font=\footnotesize]
     
    \draw[step=0.5,black,very thin,fill=gray!10] (0,0.5) grid (2.5,2.5) rectangle (0,0.5);
    
    \fill[blue!50] (0,2.5) rectangle (0.5,2.0);
    
    \end{tikzpicture}
    \caption{}
    \end{subfigure}

    \begin{subfigure}[t!]{0.25\textwidth}
    \centering
    \begin{tikzpicture}[shorten >=1pt,->,draw=black!50,font=\footnotesize]

    \fill[gray!10] (0,0.5) rectangle (2.5,2.5);
        
    \fill[red!20] (0,2.5) rectangle (0.5,0.5);
    \fill[red!50] (0,2.5) rectangle (0.5,2.0);

    \fill[blue!50] (0.5,2.5) rectangle (1,2.0);
    \draw[step=0.5,black,very thin,fill=gray!10] (0,0.5) grid (2.5,2.5);

    \end{tikzpicture}
    \caption{}
    \end{subfigure}

    \begin{subfigure}[t!]{0.25\textwidth}
    \centering
    \begin{tikzpicture}[shorten >=1pt,->,draw=black!50,font=\footnotesize]
     
    \fill[gray!10] (0,0.5) rectangle (2.5,2.5);
        
    \fill[red!20] (0,2.5) rectangle (1,0.5);
    \fill[red!50] (0,2.5) rectangle (1,2.0);

    \fill[blue!50] (1,2.5) rectangle (1.5,2.0);
    \draw[step=0.5,black,very thin,fill=gray!10] (0,0.5) grid (2.5,2.5);
    
    \end{tikzpicture}
    \caption{}
    \end{subfigure} 
     
    \begin{subfigure}[t!]{0.25\textwidth}
    \centering
    \begin{tikzpicture}[shorten >=1pt,->,draw=black!50,font=\footnotesize]
     
    \fill[gray!10] (0,0.5) rectangle (2.5,2.5);
        
    \fill[red!20] (0,2.5) rectangle (1,0.5);
    \fill[red!50] (0,2.5) rectangle (1,2.0);

    \fill[green!20] (1.5,2.5) rectangle (2.5, 2);  
    \fill[green!50] (1,2.5) rectangle (1.5, 2);  
    
    \fill[blue!50] (1,2.0) rectangle (1.5,1.5);
    
    \draw[step=0.5,black,very thin,fill=gray!10] (0,0.5) grid (2.5,2.5);
    
    \end{tikzpicture}
    \caption{}
    \end{subfigure}
    
    \\
    \\
    
    \begin{subfigure}[t!]{0.25\textwidth}
    \centering
    \begin{tikzpicture}[shorten >=1pt,->,draw=black!50,font=\footnotesize]
     
    \fill[gray!10] (0,0.5) rectangle (2.5,2.5);
        
    \fill[red!20] (0,2.5) rectangle (1,0.5);
    \fill[red!50] (0,2.5) rectangle (1,2.0);

    \fill[green!20] (1.5,2.5) rectangle (2.5, 1.5);  
    \fill[green!50] (1,2.5) rectangle (1.5, 1.5);  
    
    \fill[blue!50] (1,1.5) rectangle (1.5,1);
    
    \draw[step=0.5,black,very thin,fill=gray!10] (0,0.5) grid (2.5,2.5);
    
    \end{tikzpicture}
    \caption{}
    \end{subfigure}

    \begin{subfigure}[t!]{0.25\textwidth}
    \centering
    \begin{tikzpicture}[shorten >=1pt,->,draw=black!50,font=\footnotesize]
    \fill[gray!10] (0,0.5) rectangle (2.5,2.5);
        
    \fill[red!20] (0,2.5) rectangle (1,0.5);
    \fill[red!50] (0,2.5) rectangle (1,2.0);

    \fill[green!20] (1.5,2.5) rectangle (2.5, 1.5);  
    \fill[green!50] (1,2.5) rectangle (1.5, 1.5);  

    \fill[red!20] (1,0.5) rectangle (1.5,1);
    \fill[red!50] (1,1.5) rectangle (1.5,1);
    
    \fill[blue!50] (1.5,1.5) rectangle (2,1);
    
    \draw[step=0.5,black,very thin,fill=gray!10] (0,0.5) grid (2.5,2.5);
    
    \end{tikzpicture}
    \caption{}
    \end{subfigure}

    \begin{subfigure}[t!]{0.25\textwidth}
    \centering
    \begin{tikzpicture}[shorten >=1pt,->,draw=black!50,font=\footnotesize]
    \fill[gray!10] (0,0.5) rectangle (2.5,2.5);
        
    \fill[red!20] (0,2.5) rectangle (1,0.5);
    \fill[red!50] (0,2.5) rectangle (1,2.0);

    \fill[green!20] (1.5,2.5) rectangle (2.5, 1.5);  
    \fill[green!50] (1,2.5) rectangle (1.5, 1.5);  

    \fill[red!20] (1,0.5) rectangle (2,1);
    \fill[red!50] (1,1.5) rectangle (2,1);
    
    \fill[blue!50] (2,1.5) rectangle (2.5,1);
    
    \draw[step=0.5,black,very thin,fill=gray!10] (0,0.5) grid (2.5,2.5);
    
    \end{tikzpicture}
    \caption{}
    \end{subfigure}

    \begin{subfigure}[t!]{0.25\textwidth}
    \centering
    \begin{tikzpicture}[shorten >=1pt,->,draw=black!50,font=\footnotesize]
    \fill[gray!10] (0,0.5) rectangle (2.5,2.5);
        
    \fill[red!20] (0,2.5) rectangle (1,0.5);
    \fill[red!50] (0,2.5) rectangle (1,2.0);

    \fill[green!20] (1.5,2.5) rectangle (2.5, 1.5);  
    \fill[green!50] (1,2.5) rectangle (1.5, 1.5);  

    \fill[red!20] (1,0.5) rectangle (2.5,1);
    \fill[red!50] (1,1.5) rectangle (2.5,1);
        
    \draw[step=0.5,black,very thin,fill=gray!10] (0,0.5) grid (2.5,2.5);
    
    \end{tikzpicture}
    \caption{}
    \end{subfigure}

 \end{tabular}
 }
 \caption{Example of incremental verification algorithm's run. }
\label{fig:alg}
\end{figure}

\mysubsection{Example.} In Fig.~\ref{fig:alg}, the grid represents the options for verification queries of robustness for brightness and noise perturbation, with parameters $(\beta, \epsilon) \in [0.1,0.2,0.3,0.4,0.5]\times [0.1,0.2,0.3,0.4]$.
The purple cell represents the current tuple $(\beta, \epsilon)$. Red
marks \unsat{} queries, green marks \sat{} queries. Rich colors
represent a call for the verifier, while pale colors represent a deduction of satisfiability.
The algorithm first queries the verifier to verify robustness with parameters $(0.4, 0.1)$, which returns \unsat{}. Then, the algorithm deduces \unsat{} for queries with $\beta=0.1, \epsilon < 0.4$ without calling the verifier again, and queries the verifier to verify robustness with parameters $(0.4, 0.2)$. Since the verifier returns \unsat{} again, the algorithm deduces \unsat{} for queries with $\beta=0.2, \epsilon < 0.4$ and queries the verifier to verify robustness with parameters $(0.4, 0.3)$. This time, the verifier returns \sat{}, so the algorithm deduces \sat{} for queries with $\beta>0.2, \epsilon = 0.4$. The algorithm then queries the verifier to verify robustness with parameters $(0.3, 0.3)$. The rest of the iterations continue similarly.

\section{Evaluation}
\label{sec:eval}

\sloppy For the $1145$ correctly classified test images, we verify
$N_1$'s robustness to noise and brightness for parameters
$(\epsilon, \beta)\in [0,
0.05,0.1,0.15,0.2]\times[0,0.1,0.2,0.3,0.4,0.5]$, and to contrast
perturbations with mean pixel value $\mu = 0.2585$ and
$\gamma \in [0.1,0.2,...,0.9]$.  We make use of the incremental verification
algorithm for noise and brightness perturbations. For contrast, we run
a binary search algorithm to find the minimal $\gamma$ parameter for
which the query is \unsat{}. We use an arbitrary timeout of $22.5K$
seconds per single query, and $80$ hours for the overall runtime to
analyze a single input point. The results are summarized below and in Appendix~\ref{app:time}.  

\begin{figure}[htb]

  \begin{center}
    \vspace{-0.6cm}

    \scalebox{0.8}{
      \includegraphics[width=\linewidth, trim={0 0 4cm 1.5cm }, clip]{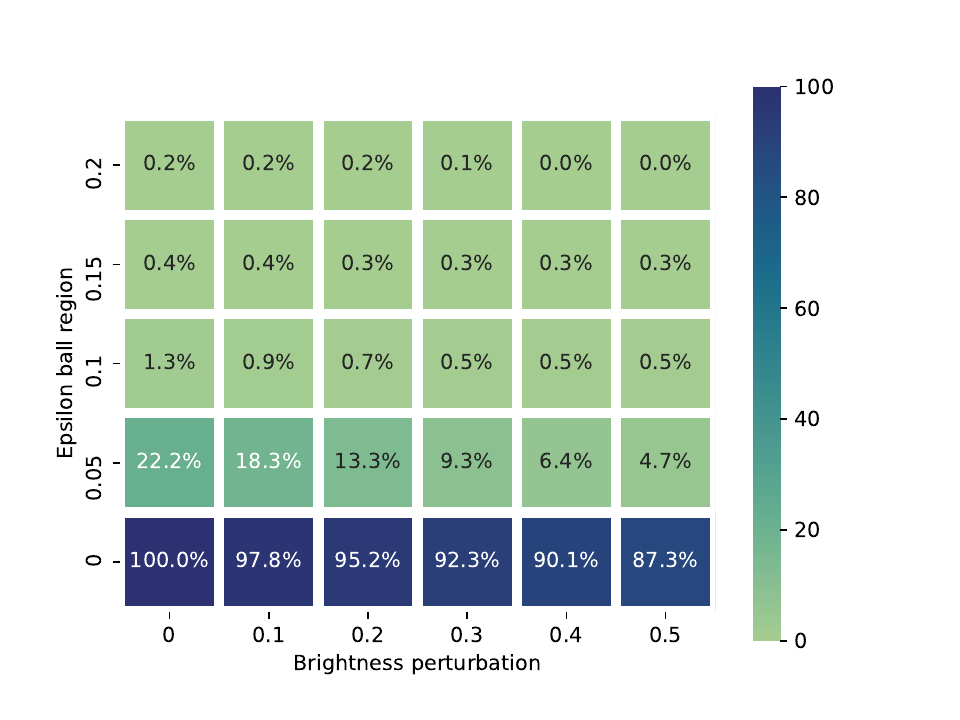}
    }
  \end{center}
   \vspace{-0.6cm}

  \caption{Percentage of \unsat{} queries per noise and brightness parameters.}
  \label{fig:brightnessheatmap}
\end{figure}
 \vspace{-0.2cm}

\begin{figure}[h]
  \begin{center}
    \scalebox{1}{
      \includegraphics[width=\linewidth, trim={0 3cm 0 5cm }, clip]{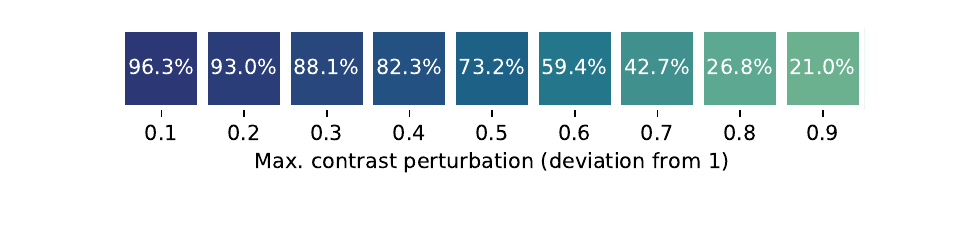}
    }
  \end{center}
    \vspace{-0.8cm}
  \caption{Percentage of \unsat{} queries per contrast parameter}
  \label{fig:contrastheatmap}
      \vspace{-0.6cm}

\end{figure}

Fig.~\ref{fig:brightnessheatmap}
shows the percentage of \unsat{} queries for noise and brightness perturbations, indicating the absence of
counter-examples, of 1097 points for which the analysis has not timed out. The
incremental verification algorithm invoked the verifier on 13231
queries, whereas the results of 59\% of the queries were deduced,
using the incremental approach, without additional invocations. 
Fig.~\ref{fig:contrastheatmap} shows the percentage of \unsat{}
queries for contrast perturbations within the range $[1-\gamma,
1+\gamma]$. The binary search algorithm invoked the verifier 3915
times, whereas the remaining 62\% of the queries were deduced without
additional invocations. For contrast perturbations, all queries
terminated without a timeout.

Overall, the results indicate that the classifier shows similar
robustness to contrast and brightness perturbations. However, it is
significantly more sensitive to noise perturbations. We note that
noise in images comes from various sources. Some noise is inherent to the
camera's sensor (e.g.,~impulse noise, thermal noise) or its associated
electronics (e.g.~shot noise). Other noise is a direct consequence of
operating and environmental conditions (e.g.,~low-light conditions,
scenery colors, etc.). Brightness and contrast also fall into this
category; they are both inherently related to operating
conditions. Although noise originating from image acquisition is
certainly a nuisance, it can in part be reduced by noise reduction
techniques, as well as an expert understanding of the camera
characteristics and continuous quality tracking. Other kinds of noise
are more challenging to predict or mitigate, as the number of different operating
conditions (weather, time of day, scenery, etc.) is effectively infinite. Therefore, it is somehow reassuring that our classifier seems to be less vulnerable to contrast and brightness, as these perturbations are highly unpredictable.


\section{Conclusion}
\label{sec:conc}
As numerous state-of-the-art image classifiers are vulnerable to small
image perturbations, robustness is a key safety requirement; and
certification authorities, such as EASA, might require confirmed
robustness as part of the model certification process in the aerospace
domain.\footnote{https://www.easa.europa.eu/en/easa-concept-paper-first-usable-guidance-level-1-
machine-learning-applications-proposed-issue-01pdf}
This work explores the challenges that the industry is facing in its
effort to safely deploy deep-learning-based systems, and the benefits that formal
methods can afford in assessing the robustness of DNN models to various perturbations. 
One significant challenge is the limited scalability of current verification
techniques.

In this work we focused on assessing the robustness of a prototype
runway object classifier provided by Airbus, with respect to three
common image perturbations types. To partially address the scalability
challenge, we  exploited the monotonicity of these perturbations in
designing an algorithm that improved the  performance of the overall
verification process. 
Moving forward, we aim to assess additional, larger, Airbus networks with  higher-resolution input; and to verify their robustness to simultaneous brightness and contrast perturbations. To improve performance, which will enable verifying larger networks, we intend to examine applying DNN abstraction methods~\cite{ElGoKa22} to the verification queries we have used. 
In addition, we aspire to increase the reliability of the results by using the \emph{proof producing} version of Marabou~\cite{IsBaZhKa22}.

\mysubsection{Acknowledgements.}
This research was partially funded by Airbus Central Research \&
Technology, AI Research and by the Israeli Smart Transportation Research Center (ISTRC).

\bibliography{main}

\newpage

{\noindent\huge{Appendix}}

\renewcommand\thesection{\Alph{section}}
\renewcommand\thesubsection{\thesection.\arabic{subsection}}
\setcounter{section}{0}

\section{Visualization of Brightness and Contrast Perturbations}
\label{app:perturbations}

\begin{figure}[h!]
\centering
\includegraphics[width=8cm]{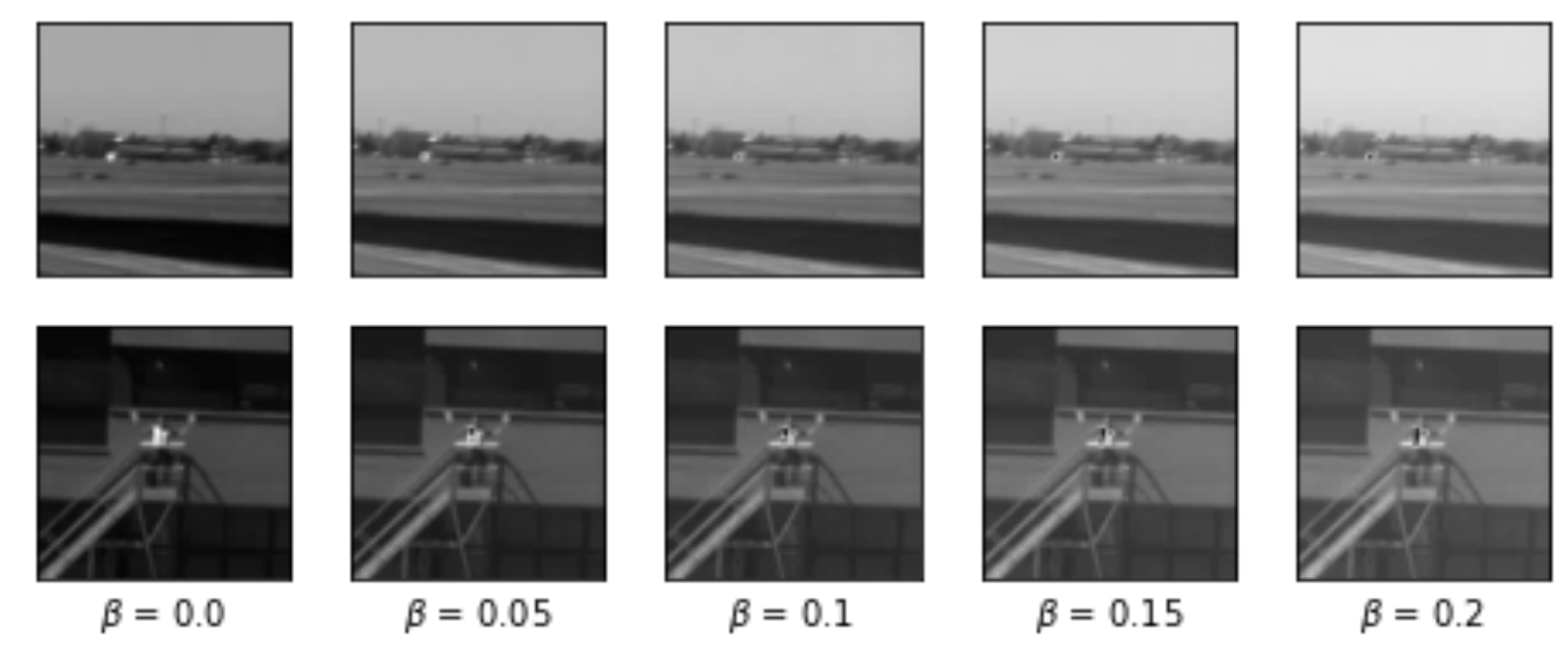}
\caption{Brightness perturbations for an `Aircraft' and a `Person' from the test set.}
\label{fig:b}
\end{figure}

\begin{figure}[h!]
\centering
\includegraphics[width=8cm]{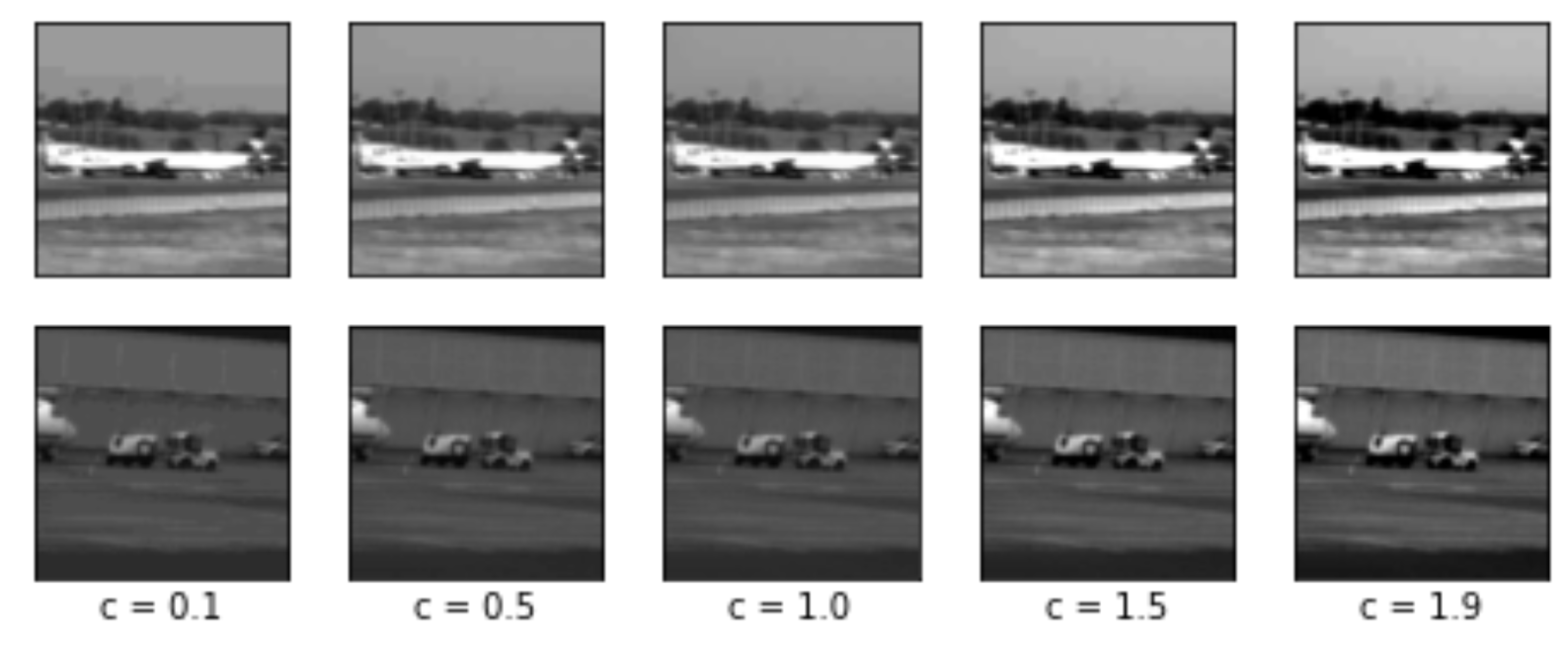}
\caption{Contrast perturbations for an `Aircraft' and a `Vehicle' from the test set.}
\label{fig:c}
\end{figure}

\begin{figure}[h!]
\centering
\includegraphics[width=8cm]{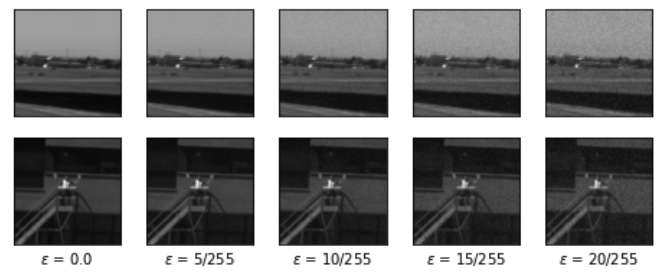}
\caption{Levels of l$_{\infty}$-norm bounded perturbations for an `Aircraft' and a `Vehicle'.}
\label{fig:n}
\end{figure}

\begin{figure}[h!]
\centering
\includegraphics[width=6cm]{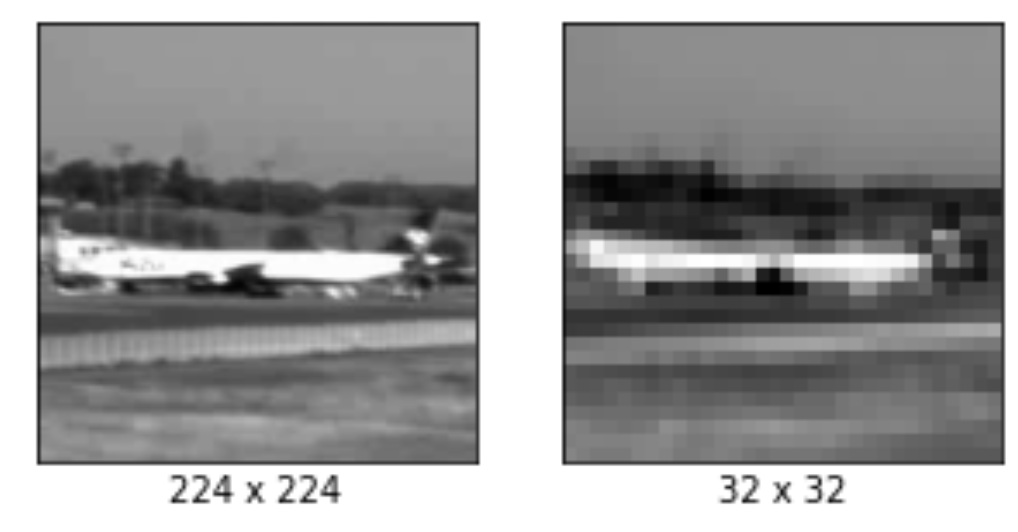}
\caption{Illustration of an `Aircraft' image at different resolutions.}
\label{fig:r}
\end{figure}

\newpage
\section{An Example of DNN}
\label{app:DNNS}
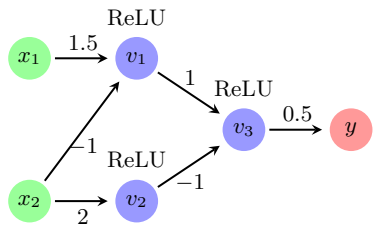
\begin{figure}[h!]
	\begin{center}
 \scalebox{0.95}{
		\def\layersep{1.5cm}
		\begin{tikzpicture}[shorten >=1pt,->,draw=black!50, node distance=\layersep,font=\footnotesize]
			
			\node[input neuron] (I-1) at (0,-1) {$x_1$};
			\node[input neuron] (I-2) at (0,-3) {$x_2$};
			
			\node[hidden neuron] (H-1) at (\layersep, -1) {$v_1$};
			\node[hidden neuron] (H-2) at (\layersep,-3) {$v_2$};
			
			\node[hidden neuron] (H-3) at (2*\layersep,-2) {$v_3$};
			
			\node[output neuron] (O-1) at (3*\layersep,-2) {$y$};
			
			\draw[nnedge] (I-1) --node[above,pos=0.5] {$1.5$} (H-1);
			\draw[nnedge] (I-2) --node[below,pos=0.5] {$-1$} (H-1);
			
			\draw[nnedge] (I-2) --node[below,pos=0.5] {$2$} (H-2);
			
			\draw[nnedge] (H-1) --node[above] {$1$} (H-3);
			\draw[nnedge] (H-2) --node[below] {$-1$} (H-3);
			
			\draw[nnedge] (H-3) --node[above] {$0.5$} (O-1);
			
			\node[above=0.05cm of H-1] (b1) {$\relu{}$};
			\node[above=0.05cm of H-2] (b1) {$\relu{}$};
			\node[above=0.05cm of H-3] (b1) {$\relu{}$};
			
		\end{tikzpicture}
  }
	\end{center}
	\caption{A toy DNN.}
	\label{fig:toyDnn}
\vspace{-1cm}
\end{figure}

Consider the DNN with $4$ layers that appears in Fig.~\ref{fig:toyDnn}, where all biases are set to zero and are ignored. For input $\langle 2,-1\rangle$, the first node in the second layer
evaluates to
$ \relu(2 \cdot 1.5 \; + \; -1 \cdot (-1) ) = \relu(4) = 4 $; 
and the second node in the second layer evaluates to
$ \relu(2 \cdot -1 ) = \relu(-2) = 0 $; Then the node
in the third layer evaluates to $ \relu(4 - 0) =4 $. and thus the output of the network is $2$. 

\section{Detailed Evaluation Results}
\label{app:time}

In this appendix, we provide a detailed description of  our experiments' results, as described in Section~\ref{sec:eval}. 
\begin{table}[h!] 
\centering
\begin{tabularx}{0.9\linewidth}{@{\extracolsep{\fill}}|c|c|c|c|c|c|c|c|c|c|}   
\hline
 Result & 0.1 & 0.2 & 0.3 & 0.4 & 0.5 & 0.6 & 0.7 & 0.8 & 0.9 \\
 \hline
 \sat{} &20.36 & 20.57 & 21.44 & 25.32 & 54.26 & 28.36 & 27.28 & 37.41 & 36.37 \\ 
  \unsat{} & 19.06 & 19.06 & 21.93 & 106.14 & 29.37 & 284.44 & 241.34 & 87.12 & 63.89 \\ 
  \unknown{}  & 0 & 0 & 0 & 0 & 0 & 0 & 51.77 & 59.21 & 55.07 \\ 
 \hline
\end{tabularx}
 \caption{Avg. solving time for contrast queries (sec.).}
 \label{table:contrasttimetable}
\end{table}

\begin{table}[h!] 
\centering

\begin{tabularx}{0.9\linewidth}{@{\extracolsep{\fill}}|c|c|c|c|c|c|c|c|}   
\hline
 \# & \backslashbox{Noise}{Bright.}  & 0 & 0.1 & 0.2 & 0.3 & 0.4 & 0.5  \\
 \hline
 \sat{} & 0 & N/A & 49.52 &   61.47 & 45.09 & 53.15 &
    66.37 \\ 
 \unsat{} & & N/A & 79.62 & 73.36 & 69.92 & 74.94 &
    83.97 \\
 \hline
 \sat{}& 0.05 &  50.99 & 38.27 & 46.42 & 373.89 & 1457.50 &
  1125.45  \\ 
 \unsat{}& & 315.12 & 423.99 & 787.40 & 1111.39 & 652.05 &
   759.83 \\
 \hline
 \sat{} & 0.1 & 51.24 & 48.28 & 34.85 & 39.99 &  422.11 &
   112.02  \\ 
 \unsat{} & & 1887.37 & 2027.04 & 159.66 & 72.11 & 89.05 &
   301.96 \\
 \hline
 \sat{} & 0.15 & 39.44 & 33.46 & 1474.75 &  745.56 & 26.68 & 22.13  \\ 
 \unsat{}& & 2359.39 & 225.52 & 84.77 & 393.32 & 259.87 &
  1026.77 \\
 \hline
 \sat{}& 0.2 & 35.63 & 18.64 & N/A & 18.61 &  N/A & N/A  \\ 
 \unsat{}& &  77.30 & 92.12 & 1165.06 & 1489.30 & N/A & N/A \\
 \hline
\end{tabularx}
 \caption{Avg. solving time for noise and brightness queries (sec.).}
 \label{table:brightnesstimetable}
\end{table}

\begin{table}[h!] 
\centering
\begin{tabularx}{0.9\linewidth}{@{\extracolsep{\fill}}|c|c|c|c|c|c|c|c|c|c|}  
\hline
 \# Queries & 0.1 & 0.2 & 0.3 & 0.4 & 0.5 & 0.6 & 0.7 & 0.8 & 0.9 \\
 \hline
 \textbf{ \sat{} Overall} & 42 & 80 & 136 & 203 & 306 & 465 & 654 & 836 & 901 \\ 
 deduced & 0 & 0 & 80 & 136 & 0 & 326 & 306 & 654 & 836 \\ 
  verified &  42 & 80 & 56 & 67 & 306 & 159 & 348 & 182 & 65  \\ 
  \hline
  \textbf{\unsat{} Overall} & 1103 & 1065 & 1009 & 942 & 839 & 680 & 490 & 308 & 241 \\ 
   deduced & 1065 & 839 & 839 & 839 & 0 & 490 & 0 & 0 & 0 \\ 
  verified & 38 & 226 & 170 & 103 & 839 & 190 & 490 & 308 & 241 \\ 
  \hline
  \textbf{\unknown{}}  & 0 & 0 & 0 & 0 & 0 & 0 & 1 & 1 & 3 \\ 
 \hline
\end{tabularx}
 \caption{Results summary for contrast queries.}
 \label{table:contrastdeducedvsverified}
\end{table}

\begin{table}[h!] 
\centering

\begin{tabularx}{0.9\linewidth}{@{\extracolsep{\fill}}|c|c|c|c|c|c|c|c|}   
\hline
 \# & \backslashbox{Noise}{Bright.}  & 0 & 0.1 & 0.2 & 0.3 & 0.4 & 0.5  \\
 \hline
 \textbf{ \sat{} Overall} & 0  & 0 & 24 & 53 & 84 & 109 & 139  \\ 
 deduced &  & 0 & 0 & 0 & 53 & 53 & 109  \\ 
  verified &  & 0 & 24 & 53 & 31 & 56 & 30  \\ 
  \textbf{\unsat{} Overall} &  & 1097 & 1073 & 1044 & 1013 & 988 & 958 \\ 
   deduced &  & 1097 & 1044 & 235 & 889 & 67 & 51 \\ 
  verified &  & 0 & 29 & 809 & 124 & 921 & 907 \\ 
  \hline
 \textbf{ \sat{} Overall} &  0.05 & 325 & 371 & 424 & 484 & 533 & 1093  \\ 
 deduced &  &  0 & 325 & 371 & 325 & 483 & 533  \\ 
  verified &  & 325 & 46 & 53 & 159 & 50 & 560  \\ 
  \textbf{\unsat{} Overall} &  & 235 & 194 & 146 & 100 & 69 & 52 \\ 
   deduced &  & 14 & 10 & 8 & 6 & 6 & 3    \\ 
  verified &  & 221 & 184 & 136 & 94 & 63 & 49 \\ 
  \hline
 \textbf{ \sat{} Overall} &  0.1 & 829 & 859 & 884 & 900 & 934 & 936  \\ 
 deduced &  & 0 & 829 & 859 & 829 & 900 & 934 \\ 
  verified &  &  829 & 30 & 25 & 71 & 34 & 2  \\ 
  \textbf{\unsat{} Overall} &  & 14 & 10 & 8 & 6 & 6 & 6 \\ 
   deduced &  & 4 & 4 & 3 & 3 & 3  & 0 \\ 
  verified &  & 10 & 6 & 5 & 3 & 3 & 6 \\ 
  \hline
 \textbf{ \sat{} Overall} & 0.15 & 1036 & 1045 & 1054 & 1064 & 1070 & 1075  \\ 
  deduced &  & 0 & 1036 & 1047 & 1036 & 1064 & 1070  \\ 
  verified &  & 1036 & 11 & 7 & 28 & 6 & 5 \\ 

  \textbf{\unsat{} Overall} &  & 4 & 4 & 3 & 3 & 3 & 3 \\ 
   deduced &  & 2 & 1 & 1 & 1 & 0 & 0 \\ 
  verified &  &  2 &  3 & 2 & 2 & 3 & 3 \\ 
   \hline
 \textbf{ \sat{} Overall} & 0.2 & 1087 & 1091 & 1091 & 1091 & 1091 & 1091  \\ 
 deduced &  &  0 & 1087 & 1091 & 1087 & 1091 & 1091  \\ 
  verified &  &1087 & 4 & 0 & 4 & 0 & 0 \\ 
  \textbf{\unsat{} Overall} &  & 2 & 1 & 2 & 1 & 0 & 0 \\ 
   deduced &  & 0 & 0 & 0 & 0 & 0 & 0 \\ 
  verified &  &  2 & 1 & 2 & 1 & 0 & 0 \\ 
  \hline
\end{tabularx}
 \caption{Summary of results for noise and brightness queries (excluding timeouts).}
 \label{table:brightnessdeducedvsverified}
\end{table}

\newpage
\section{The Network is more Robust to Brightness Perturbations than to Noise}
\label{app:brightnessRobust}

\begin{figure}[h]
\centering
\includegraphics[width=10cm]{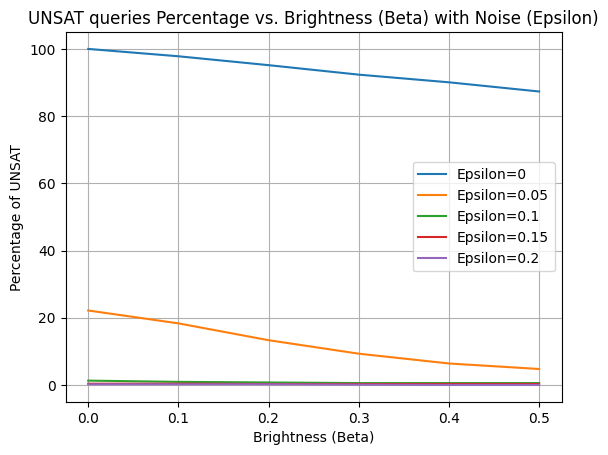}
\caption{Percentage of queries with \unsat{} answer with respect to different brightness perturbations for every noise perturbation.}
\label{fig:beta_x}
\end{figure}

\begin{figure}[h]
\centering
\includegraphics[width=10cm]{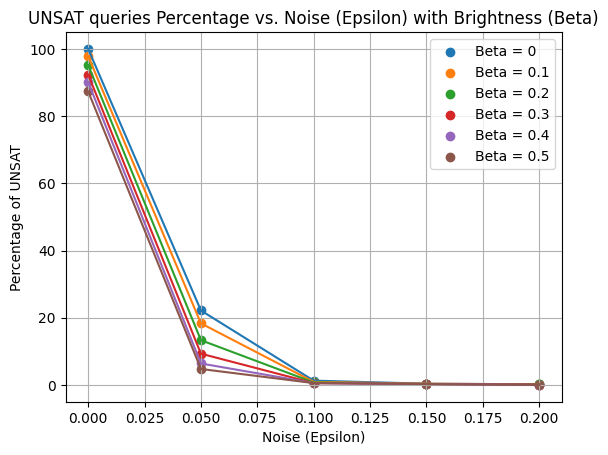}
\caption{Percentage of queries with \unsat{} answer with respect to different noise perturbations for every brightness perturbation.}
\label{fig:epsilon_x}
\end{figure}

In both Fig. \ref{fig:beta_x} and Fig. \ref{fig:epsilon_x}, we show the percentage of the verification queries that had an \unsat{} result with respect to certain brightness and noise perturbations. In Fig.~\ref{fig:beta_x}, we notice that changes in the brightness perturbation parameter $\beta$ have a minor effect on the queries' answers. On the other hand, as emphasized in  Fig. \ref{fig:epsilon_x}, changes in the noise perturbation parameter dramatically decrease the network's performance. This observation tells us that the network is more robust and handles brightness perturbations better than noise.

\end{document}